# HGO-YOLO: Advancing Anomaly Behavior Detection with Hierarchical Features and Lightweight Optimized Detection

Qizhi Zheng, Zhongze Luo, Meiyan Guo, Xinzhu Wang, Renqimuge Wu, Qiu Meng, Guanghui Dong*

College of computer and control engineering, Northeast Forestry University, Harbin, 150040, China

**Abstract –**

**Accurate, real-time object detection on resource-constrained hardware is critical for anomaly-behavior monitoring. We introduce HGO-YOLO, a lightweight detector that combines GhostHGNetv2 with an optimized parameter-sharing head (OptiConvDetect) to deliver an outstanding accuracy–efficiency trade-off. By embedding GhostConv into the HGNetv2 backbone with multi-scale residual fusion, the receptive field is enlarged while redundant computation is reduced by 50 %. OptiConvDetect shares a partial-convolution layer for the classification and regression branches, cutting detection-head FLOPs by 41 % without accuracy loss. On three anomaly datasets (fall, fight, smoke), HGO-YOLO attains 87.4 % mAP@0.5 and 81.1 % recall at 56 FPS on a single CPU with just 4.3 GFLOPs and 4.6 MB—surpassing YOLOv8n by +3.0 % mAP, −51.7 % FLOPs, and 1.7 × speed. Real-world tests on a Jetson Orin Nano further confirm a stable throughput gain of 42 FPS.**

**Keywords - YOLOv8, HGNetv2, Lightweight, Anomaly Behavior Detection**

## 1. **Introduction**

In contemporary society, the detection of anomalous behavior in both public and private buildings has become pervasive, especially with the widespread deployment of cameras and video surveillance systems. These systems are engineered to monitor for abnormal behavior to identify potential threats or safety hazards, which is a crucial aspect of maintaining public safety. However, traditional surveillance systems depend a lot on human operators. They need these operators to watch video feeds. This can lead to inefficiencies because of human traits. These traits include fatigue and lapses in concentration. These factors often lead to missed events or slow responses. This can make it hard to detect potential threats on time. It also reduces the effectiveness of keeping public safety and peace.

Intelligent detection systems have been developed to meet this challenge. These systems can automatically find different types of abnormal behavior. This paper will explore three common situations for detecting abnormal behavior. These situations are falling, fighting, and smoking[1, 2]. By helping to quickly and accurately identify abnormal behavior, these systems offer strong tools. They can be used in areas like surveillance, security, healthcare, and more. This effectively addresses potential threats. By automating the identification process, these systems lessen the workload on human operators. They also improve the speed and accuracy of surveillance systems. This helps with better monitoring and management of abnormal behavior. The use of this technology has great potential. It can improve public safety and system efficiency in many sectors[3].

The rise of deep learning technology has greatly expanded research opportunities. Deep learning methods can automatically find and understand high-level features in complex data. This ability helps in detecting unusual behavior. OpenPose[4] is a technology that focuses on human pose estimation. It provides detailed information about human movements by detecting many key points. OpenPose is very good at monitoring changes in posture. It is especially useful for finding abnormal behaviors like falls. This is because it can capture small movements in posture. However, OpenPose is quite complex. It requires a lot of hardware resources and computing power. It may not work well in difficult situations, such as when something blocks the view. In contrast, LSTM (Long Short-Term Memory) networks[5] are a type of recurrent neural network. They are designed for working with sequence data. These networks have been widely used to model how abnormal behavior changes over time. By using their long-term memory units, LSTMs can keep important long-term information. They also perform well in capturing complex patterns in time-series data.

However, LSTMs can face challenges. These challenges include gradient vanishing or explosion. This often happens when they work in long sequences. As a result, their performance may not be very good in situations where abnormal behavior changes quickly[6]. On the other hand, the YOLO (You Only Look Once) algorithm provides efficient object detection[7]. It works by dividing the image into grids. Then, it predicts targets at the same time. This allows for real-time processing and a global view of the image. YOLO is strong in its speed and adaptability. This enables it to perform very well in real-time surveillance applications. However, it can face challenges. This happens when it has lower localization accuracy. This issue often occurs with small or closely packed targets. The YOLO algorithm was chosen for the goals and requirements of this research. Its ability to work in real time and its efficiency match the need for quick monitoring of abnormal behavior. Furthermore, its global perception feature helps it adapt to different situations and complex environments. Finally, its simple network structure helps reduce the cost and time needed for system development. After thinking carefully about these factors, we chose to use the YOLO algorithm as the main framework for detecting abnormal behavior in this study. Our proposed model, HGO-YOLO, is based on YOLOv8. It aims to improve the performance and accuracy of detecting abnormal behavior.

This paper shows the main contributions. These contributions are as follows:

1) The HGNetv2 architecture is utilized for hierarchical feature extraction within the backbone network, with the goal of enhancing its capability to process complex image data. This architecture incorporates multiple convolutional layers with diverse filter sizes, allowing the network to capture a broad spectrum of features and thus improving its overall performance.

2) Replacing Conv convolution with GhostConv not only effectively reduces the number of computations and parameters but also excels in efficiently extracting feature information.
3) Design an OptiConvDetect detection head that not only leverages the superior performance of decoupled heads but also reduces model parameters and computational costs, making the model more lightweight and efficient.

Current lightweight detectors face two key challenges: (i) conventional down-sampling and pruning degrade multi-scale features, reducing accuracy on small objects, and (ii) independently designed decoupled heads introduce redundant features. To overcome these issues, HGO-YOLO proposes an innovative hierarchical feature–co-optimization framework that: (1) preserves feature diversity through HGNetv2's stacked multi-scale convolutions; (2) employs GhostConv for efficient feature mapping; and (3) introduces the OptiConvDetect head to enable cross-task parameter sharing. Experiments demonstrate that the framework sustains real-time performance at 56 FPS while achieving 87.4 % mAP@0.5, a 3 % improvement over the baseline, thereby offering a new solution for lightweight object detection.

The paper is structured as follows: Section 2 introduces related work, with a particular focus on state-of-the-art methods in recognizing human abnormal behavior. Section 3 provides a detailed description of our proposed HGO-YOLO. Section 4 validates the model's effectiveness through experiments. Finally, Section 5 presents the conclusion.

## 2. **Related Work**

To address the challenges of multiscale targets and complex image backgrounds, researchers have proposed various improved algorithms[8]. To address the issue of multi-scale object detection, some researchers have focused on reconstructing the feature pyramid structure. For instance, Liu et al. [9]developed the PDT-YOLO algorithm based on YOLOv7-tiny. They integrated the PANet structure into the head component, incorporating bottom-up augmentation routing and adaptive feature pooling operations. This method allows interaction and fusion between features at different levels. It improves detection accuracy for objects of different sizes. In addition, in the study[10], QAFPN was suggested to fully combine feature information from four layers of different sizes. This stops information from being lost or becoming less useful. It also reduces the difference between layers that are not next to each other. As a result, the model becomes better at detecting objects of various sizes. These studies improve detection accuracy through cross-level feature fusion and adaptive feature pooling. They also make the feature maps more expressive and help the model detect objects more efficiently. However, these improvements come with some downsides. The model becomes more complex, which leads to higher computational costs during training and inference.

To meet the needs for real-time performance, researchers often use lightweight backbone networks. Cui et al.[11] combined MobileNet with the YOLOv4 framework. This created a lightweight backbone for feature extraction. They designed multi-scale fusion modules. These modules add more feature information. As a result, they created a lightweight object recognition model. This model supports fast and accurate detection of targets in different situations. In a similar way, [12]introduced LFF-YOLO. It is based on YOLOv3 and uses ShuffleNetv2 as the backbone. This approach achieved big reductions in parameters and inference speed. However, these lightweight networks have a limited number of layers and parameters. This can limit their ability to capture complex features and details. As a result, it may lead to not representing intricate patterns well enough.

Additionally, some researchers use lightweight convolutions. This helps to decrease the number of parameters. Qin et al. [13]added depth-separable convolution to YOLOv3. This change significantly reduces the number of parameters and the amount of computation. Importantly, it does this without losing accuracy. This allows the effective use of deeper and wider neural networks. Similarly, [14]introduced the DualConv lightweight deep neural network, which reduced the number of parameters of MobileNetV2 by 54%. Zhao et al.[15] used ShuffleNetv2 as the backbone for extracting features from pomegranates. They utilized group convolution to reduce computational load and increased channel interaction through random channel ordering. These lightweight techniques lower computation and parameter counts, enhancing processing speed and efficiency, but they may also decrease accuracy and limit feature extraction capabilities.

In recent years, many efficient visual architectures have been proposed. EfficientViT, based on Vision Transformer, improves computational efficiency by optimizing the self-attention mechanism, making it suitable for resource-constrained devices[16]. VanillaNet simplifies the network structure and reduces complexity to improve computational efficiency, but it may not perform as well as other models in complex tasks[17]. FasterNet focuses on accelerating object detection, with an advantage in real-time performance[18]. GhostNet uses Ghost convolution technology to reduce computation and memory usage, making it suitable for mobile devices[19]. Mobilenetv4 optimizes computation through depthwise separable convolutions and efficient design, making it especially suitable for low-power devices[20]. Swin Transformer introduces a sliding window self-attention mechanism that effectively captures multi-scale features, improving performance in tasks like image classification and object detection[21]. These models contribute to achieving efficiency and lower computational complexity and are widely used in various visual tasks. In contrast, HGO-YOLO combines the multi-scale feature extraction capability of HGNetv2 with the lightweight design of GhostConv, achieving a better balance between accuracy and efficiency.

## 3. **Methods**

### 3.1 **YOLOv8 Algorithm**

The YOLO series algorithms exhibit a high degree of similarity. The YOLOv8 model used in this study is primarily composed of three components: the backbone, neck, and head, as shown in Fig 1.

The backbone network is tasked with extracting feature information from the input image. YOLOv8 adopts the C2f (Cross Stage Partial Network Bottleneck with 2 Convolution) structure to enhance feature extraction efficiency and facilitate richer gradient flow information, thereby improving the understanding of image content while maintaining a lightweight design. Additionally, the SPP (Spatial Pyramid Pooling) mechanism is enhanced to SPPF through serial and parallel modifications. YOLOv8's neck network adopts a similar architecture to FPN (Feature Pyramid Network) used in PANet, effectively integrating features of different scales, particularly excelling in multi-scale object detection tasks. The YOLOv8 head network is responsible for detecting targets and providing the coordinates and categories of the detection box. To achieve this, it employs the Decoupled-Head structure, which separates the regression and prediction branches. Additionally, it integrates the integral form representation from the Distribution Focal Loss strategy to manage the regression branch. This strategic approach enhances target detection performance by transforming coordinate predictions from a deterministic single value to a distribution.

### 3.2 **HGO-YOLO**

To enhance accuracy and facilitate rapid recognition of abnormal behaviors, we introduce the HGO-YOLO method, whose architecture is depicted in Fig 1. The backbone structure consists of HGStem, HG_Block, DWConv, Ghost_HGBlock, and SPPF modules. Regarding the head network, we propose a novel detection head named OptiConvDetect, which reduces model parameters and computational costs by employing parameter sharing and PConv (Partial Convolution). HGO-YOLO uses the MPDIoU loss function, enhancing its performance.

#### 3.2.1 HGNetv2

The HGNetv2 network captures multi-level information in graph structures by extracting features and aggregating information at different scales.

It utilizes graph convolution operations to aggregate relationships between nodes, adaptively adjusts information aggregation strategies, and enhances feature representation capability and model robustness. The structure diagram of HGNetv2 is shown in Fig 1.

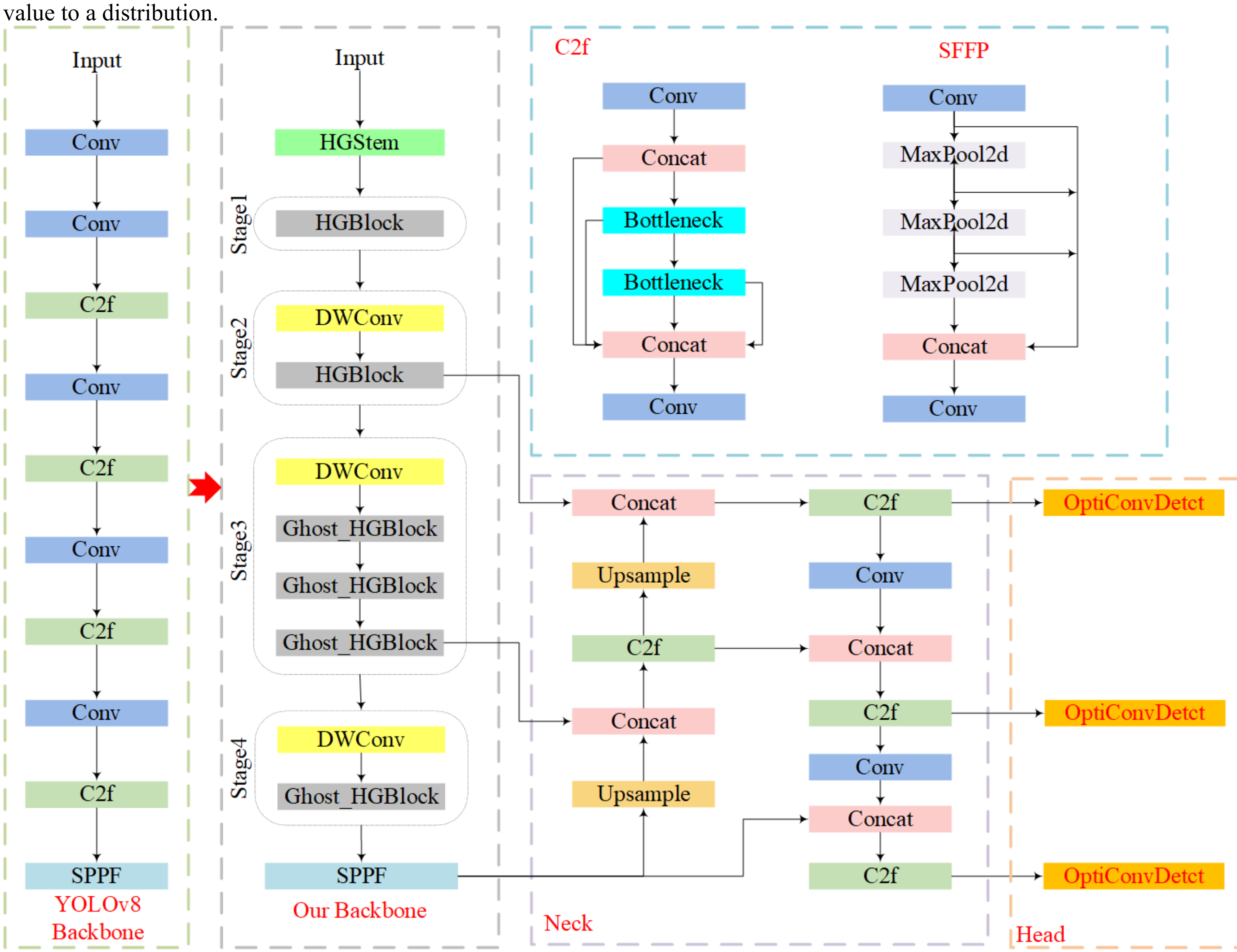

Fig. 1. Overall architecture of HGO-YOLO and its integration into YOLOv8. Step ①: HGNetv2 replaces the original YOLOv8 backbone, preserving multi-scale features via stacked hierarchical convolutions; Step ②: GhostConv substitutes the standard convolution in each HG_Block to efficiently propagate redundant information; Step ③: the fused feature maps are then passed to the detection head for classification and bounding-box regression.

As shown in Fig 1, the structure of HGNetv2 is composed of multiple HGBlock. The specific structure of the HGBlock module is illustrated in Fig 2. The HGBlock module enhances feature representation capabilities through multiple layers of convolution, feature concatenation, channel compression and expansion, and residual connections, with the ability to handle multi-scale features. The module starts with an input feature map $X$ of dimensions $c \times h \times w$. By stacking $Layer_num$ convolutional blocks, each consisting of a convolutional layer, batch normalization (BN), and an activation function (ReLU), the module captures features at different scales due to varying receptive fields of these convolutional blocks. The outputs of these convolutional blocks are denoted as $Y_0 \sim Y_{Layer_num-1}$. After passing through all convolutional layers, the module concatenates the output feature maps along the channel dimension to preserve multi-scale feature information, as described by the following formula:

$$Z = Concat(Y_0, Y_1, ..., Y_{Layer_num-1}) \quad (1)$$

The concatenated feature map has dimensions $(Layer_num \times c_{mid}) \times h' \times w'$, where $c_i$ is the number of outputs channels of each convolutional block, and $h'$ and $w'$ are the height and width after the convolution operations. This concatenated feature map is then passed through a convolutional layer to compress the number of channels to out_channels/2. This step reduces computational complexity and enhances feature compactness, resulting in the feature map $Z'$.

This step reduces computational complexity and enhances feature compactness, resulting in the feature map $Z''$.This adjustment in the number of channels facilitates feature fusion and channel adaptation. By using a shortcut connection, the aggregated output is combined with the original input $X$, ensuring the retention of original information within the network to mitigate information loss. Finally, the entire module undergoes batch normalization and activation to further improve training stability and model generalization.

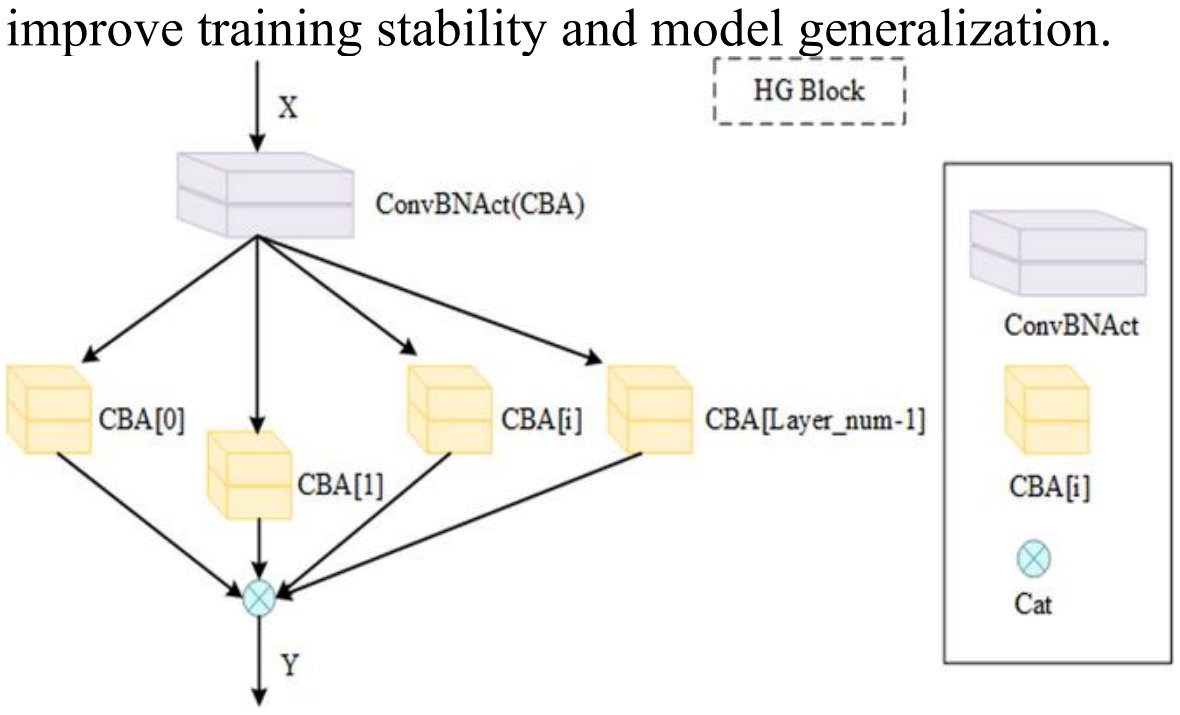

Fig 2: HG_Block module structure diagram.

3.2.2 Ghost_HGBlock

To reduce the model's size and computational load while maintaining performance, I prioritized lightweight solutions. Han et al.[21] extensively explored the information within feature maps and introduced a more lightweight Ghost convolution (GhostConv) module by fully leveraging feature map redundancy. The main difference between traditional convolution (Conv) and GhostConv lies in the number of parameters and computational efficiency, as depicted in Fig 3, providing a clear illustration of their operations. Traditional convolution, as shown in Fig 3(a), takes an input feature map $X \in R^{c \times h \times w}$ where $c$ represents the number of input channels, and $h$ and $w$ are the input height and width, respectively. The convolution operation is mathematically described by Eq (2), yielding an output n-dimensional feature map $Y \in R^{h' \times w' \times n}$ ,Here, the convolution kernel $Y \in R^{c \times k \times k \times n}$ ( with $k$ as the size of the convolution kernel), and $b$ signifies the bias term. Therefore, the computational cost of this layer is denoted by $FLOP_{S1}$ as shown in Eq (3). In traditional convolution, each input channel has corresponding convolutional kernel weights and biases, and typically, $n$ and $c$ are large, resulting in a relatively large number of parameters and computational load.

$$Y = X * f + b \quad (2)$$

$$FLOP_{S1} = n \times h' \times w' \times c \times k \times k \quad (3)$$

The operation flow of the Ghost module consists of three steps: Firstly, the input image is convoluted once by halving the number of channels and using Conv with a convolution kernel size of $1 \times 1$ , integrating features to generate a condensed feature map of the input feature layer. Secondly, a $3 \times 3$ convolution kernel is applied to the condensed feature map obtained from the previous step. This step is aimed at performing a cost-effective operation, resulting in a redundant feature map known as cheap operations. Finally, the condensed feature map and the redundant feature map generated in the first two steps are concatenated together. This concatenation process produces the final output feature map. For a visual representation of this structure, please refer to Fig 3(b).

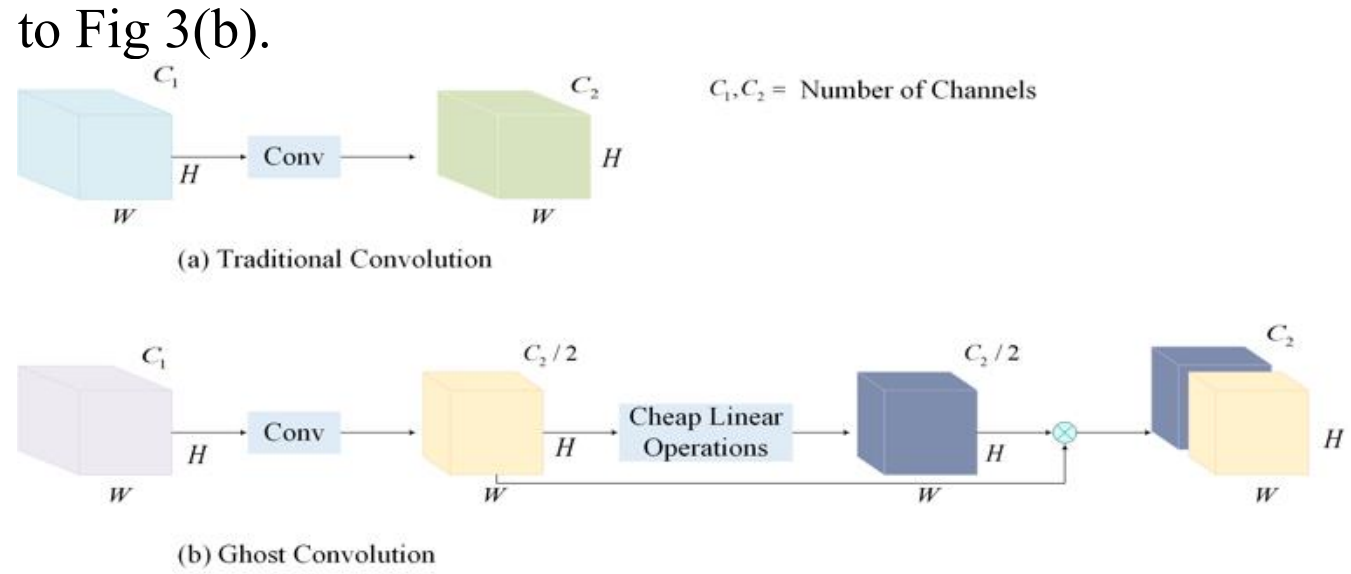

Fig 3: Ghost convolution vs. standard convolution.

The Ghost module operates by taking an input feature map $X_1 \in R^{c \times h \times w}$ and producing an output feature map $Y_1 \in R^{h' \times w' \times n}$ using a convolution kernel $f_1 \in R^{c \times k \times k \times n}$ (where $m < n$ ). For simplicity, the bias term $b$ is omitted. The output $Y_1$ then undergoes $s-1$ group convolution operations with a kernel size of $d \times d$ , which is customizable. This results in $s-1$ feature maps of size $m \times h' \times w'$ . Finally, these feature maps are added to $Y_1$ to obtain a feature map of size $n \times h' \times w'$ . The computational cost of the Ghost module, denoted as $FLOP_{S2}$ , can be calculated using Eq (4).

$$FLOP_{S2} = \frac{n}{s} \times h' \times w' \times c \times k \times k + (s-1) \times h' \times w' \times \frac{n}{s} \times d \times d \quad (4)$$

The ratio of computational cost between the two approaches can be expressed using Eq (5).

$$\frac{FLOP_{S1}}{FLOP_{S2}}=\frac{c\times k\times k}{\frac{1}{s}\times c\times k\times k+\frac{(s-1)}{s}\times d\times d}\approx\frac{s\times c}{s+c-1}=\frac{s}{1+\frac{s-1}{c}}\approx s \quad (5)$$

It can be observed that the computational cost of GhostConv is reduced to $1/s$ of the original cost when compared to traditional convolution.

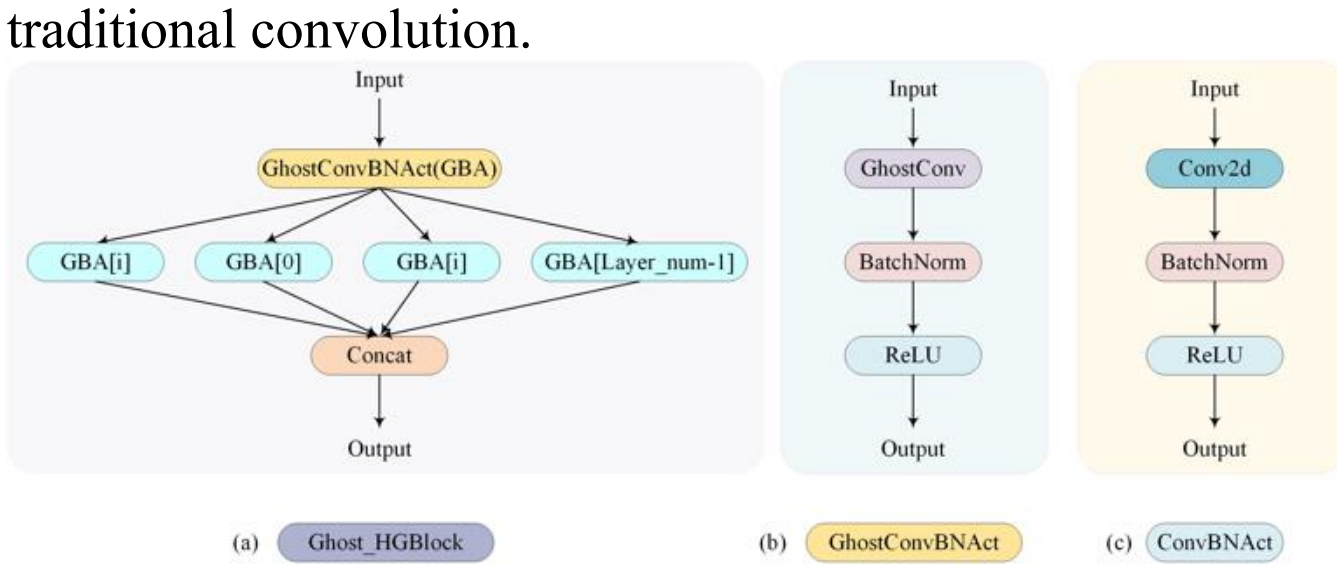

Fig 4: (a) The internal structure of the Ghost_HGBlock module. (b) Ghost Convolution with Batch Normalization and Activation (orange in Ghost_HGBlock). (c) Convolution with Batch Normalization and Activation.

Grounded in information-bottleneck theory, HGO-YOLO innovatively couples hierarchical feature preservation (HGNetv2) with redundancy-aware optimization (GhostConv). HGNetv2 employs cross-layer connections and dynamic receptive-field fusion to markedly enhance feature expressiveness in complex scenes, while GhostConv derives ghost feature maps under sparsity constraints, trimming parameters without over-compressing information. Experiments confirm that this synergy boosts detection accuracy for critical classes (e.g., “person” and “fight”) by 3.0 % while cutting overall computation by 51.7 % (Table 3), substantiating the framework’s theoretical advantage in simultaneous feature compression and information retention.

### 3.2.3 OptiConvDetect

To improve the accuracy of predicting object classes and locations, YOLOv8 employs a decoupled detection head module that separates the classification and regression branches, as shown in Fig. 5. This module divides the object detection task into two main branches: one for object classification and the other for location regression. During detection, spatial feature extraction starts with two convolutional layers, followed by processing the output channels with final convolutional layers to generate prediction information. Assuming the input feature map is $X$, after processing through two convolutional layers, the resulting feature map is, which is then processed through the regression branch $R$ and classification branch $C$ to produce the final predictions:

$$R(X')=Conv_{reg}(X') \quad (6)$$

$$C(X')=Conv_{cls}(X') \quad (7)$$

However, integrating three detection heads leads to a total of 12 $3\times3$ convolutional layers and 6 $1\times1$ convolutional layers, significantly increasing the model parameters, making the detection head account for 41.4% of YOLOv8’s GFLOPs.

To address this challenge, we propose an optimized model structure by introducing parameter sharing in the detection head, called OptiConvDetect. Unlike traditional methods, OptiConvDetect achieves parameter sharing by introducing a single PConv[22] convolutional layer $P$, instead of separate regression and classification convolutional layers. This strategy avoids the separate handling of regression and classification branches as done in conventional methods, significantly reducing redundant computations and parameter counts. Specifically, OptiConvDetect uses parameter sharing to replace multiple convolutional layers in YOLOv8’s detection head, making the detection head structure more compact, improving computational efficiency, and reducing model complexity.

The optimized structure achieves parameter sharing by introducing a single PConv convolutional layer $P$, followed byan additional convolutional layer $Conv_{opt}$ for the output channels, as shown below:

$$P(X)=PConv(X) \quad (8)$$

$$Conv_{opt}(P(X))=Conv(P(X)) \quad (9)$$

This optimization avoids using separate convolutional layers for each branch (regression and classification), which significantly reduces redundant computations and parameters, while maintaining the prediction accuracy.

In Fig. 5, we introduce an improved convolutional layer design in the detection head module to reduce model complexity by sharing convolutional layer parameters. The feature extraction module consists of PConv and $1\times1$ convolutional layers, which are reused across different detection heads to achieve parameter sharing. These feature extraction modules can be reused on different input feature maps, further facilitating parameter sharing. For regression prediction, assuming the feature map $X$ is processed through shared PConv and $1\times1$ convolutional layers to obtain $X''$, the regression prediction $\hat{B}$ is generated as follows:

$$B=DFL(Conv_{reg}(X'')) \quad (10)$$

Similarly, the classification prediction $\hat{C}$ is obtained in the same manner:

$$C=Conv_{cls}(X'') \quad (11)$$

During forward propagation, the parameter sharing mechanism for feature extraction and prediction generation is reused across each detection head. Additionally, the model dynamically computes and updates anchors $A$ and strides $S$ based on the input shape, allowing the model to share and adjust parameters under different input conditions, thereby enhancing model flexibility and efficiency:

$$A,S=DynamicAnchorStride(X,shape) \quad (12)$$

This design enables the model to automatically adjust parameters while handling different input sizes, maintaining efficient computation and accurate predictions.

## 4. **Experimental Analysis and Visualization**

### 4.1 **Dataset Description**

The dataset utilized in this study is composed of six publicly available sources, detailed as follows:

- Fall dataset: This includes the UR Fall Detection Dataset, Fall Detection Dataset, Multiple Cameras Fall Dataset, and selected segments of the COCO Dataset.
- Fight dataset: This consists of the Surveillance Camera Fight Dataset, A Dataset for Automatic Violence

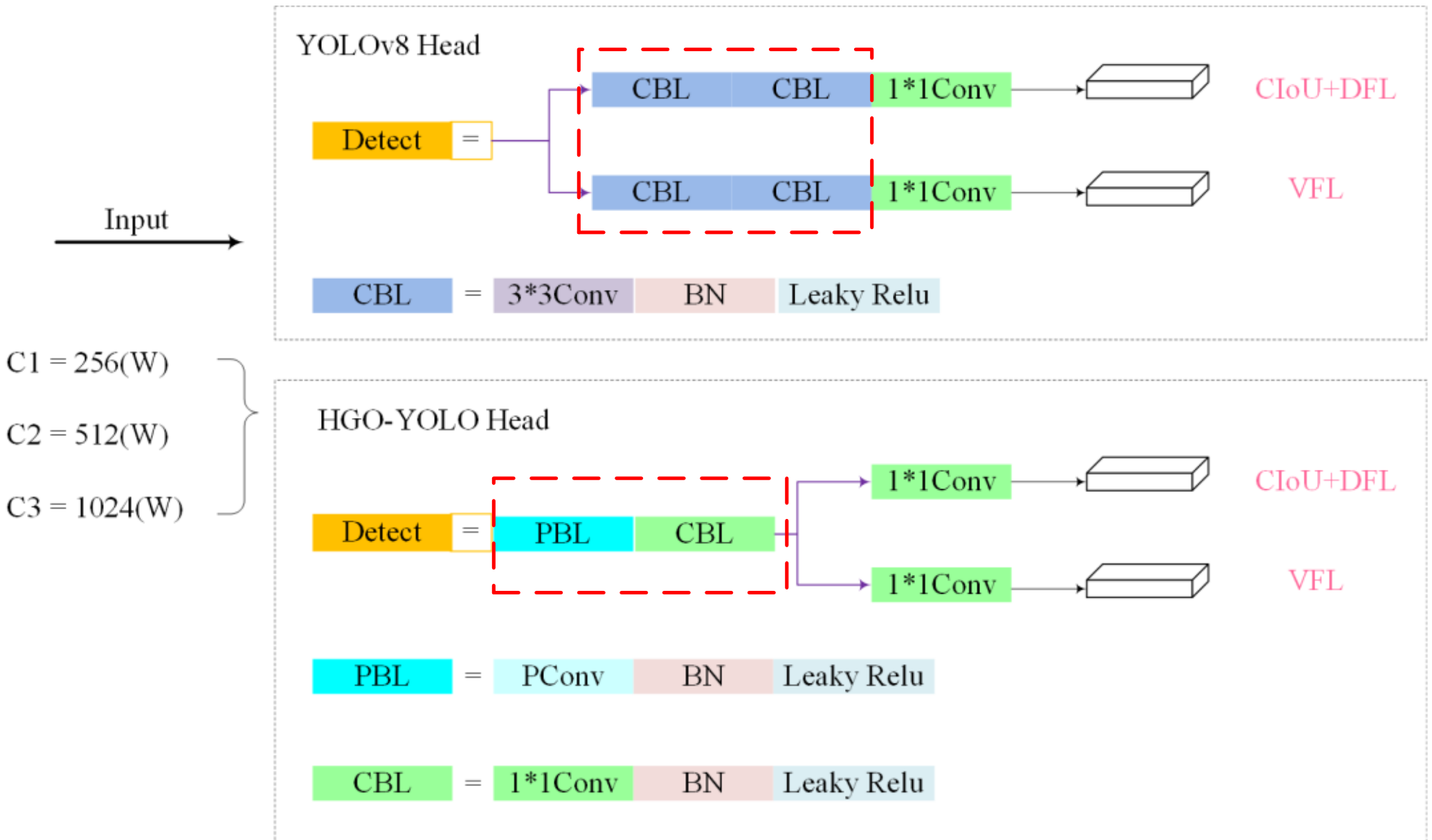

Fig 5: YOLOv8 Head vs. HGO-YOLO Head (The parameter-sharing locations are indicated by red dashed boxes.)

Detection in Videos, and the Real-Life Violence Situations Dataset.

- Smok dataset: This subset comprises 2,912 images collected from real-world scenarios on the Internet.

From the raw video streams in the public datasets, we first extracted every frame to obtain still images. To reduce temporal redundancy while preserving informational integrity, we applied temporal subsampling at 0.5 s intervals (i.e., one frame every 0.5 s); manual inspection confirmed that this rate retains more than 90 % of key action frames. The resulting corpus comprises 3,224 fight images and 4,065 fall images, for a total of 10,201 samples. During annotation, we followed rigorous guidelines and assigned each image to one of four categories—fall, fight, smoke, or person—with class counts reported in Table 1. The dataset was then partitioned into training, validation, and test subsets in an 8:1:1 ratio using stratified sampling to ensure balanced class proportions across splits. Model training employed several data-augmentation strategies, including random horizontal flipping ($p = 0.5$), brightness perturbation ($\Delta = 0.4$), and MixUp ($\lambda = 0.2$). To address the scarcity of smoke samples, we additionally introduced local occlusion augmentation, thereby increasing sample diversity and enhancing the model's robustness to minority classes.

**Table 1**: Summary of Class Distribution.

| **Class** | **Total** |
|---|---|
| Fall | 4252 |
| Fight | 4202 |
| Smoke | 3009 |

## 4.2 Experimental Setup

To comprehensively evaluate the performance of the HGO-YOLO algorithm, we implemented and conducted comparative analysis of several algorithms based on the PyTorch framework, including SSD, Faster R-CNN, the basic versions of the YOLO series (v5, v6, v7, v8, v9, v10, v11), and the RT-DTETR[23] algorithm, with the improved YOLOv8n serving as the benchmark. The experimental platform utilized an Intel(R) Xeon(R) Silver 4310 CPU @ 2.10GHz, an NVIDIA A100 80GB PCIe GPU, and the Ubuntu 20.04.2 operating system. The detailed hyper-parameter settings employed during model training are summarized in Table 2. For testing FPS on the CPU, the input resolution was set to 640×640 with a batch size of 1, and no FP16/INT8 optimizations were applied.

**Table 2**: Hyper-parameter Settings for Model Training.

| **Hyperparameter** | **value** |
|---|---|
| Input Resolution | 640×640 |
| Batch Size | 32 |
| Epochs | 200 |
| Optimizer | AdamW |
| Learning Rate | 0.001 |

## 4.3 Evaluation Indicators

To fairly evaluate how well the HGO-YOLO model works, this study uses common measures. These are mAP, FPS, and FLOPs. Mean Average Precision (mAP) is an important measure. It helps to check how well object detection models work, especially with many classes. It gives a full evaluation of how well the model can find objects in different classes. It does this by calculating the Average Precision (AP) for each class. The calculation formula is given in (13). AP comes from the precision-recall curve (PR curve) that the model creates. This curve shows the balance between precision (the ratio of true positive results to all positive results) and recall (the ratio of true positive results to all actual positive cases). The confidence threshold of the model changes as this balance is shown. The formula for AP is below:

$$Precision = \frac{TP}{TP + FP} \tag{12}$$

$$Recall = \frac{TP}{TP + FN} \tag{13}$$

$$AP = \sum_{n} \left( R_n - R_{n-1} \right) P_n \tag{14}$$

TP stands for True Positives. It is the number of cases that the model correctly classified as positive. FP stands for False Positives. It is the number of cases where the model incorrectly said something was positive, but it was negative. FN (False Negatives) represents the number of instances incorrectly classified as negative by the model when, they are positive.

FLOPs represent the computational load during the inference phase. Lower FLOPs indicate faster inference and reduce computational requirements, making the model more efficient, especially in scenarios where real-time recognition or processing is necessary.

4.4 **Comparison of Different Backbone Networks**

As shown in Table 3, the comparative experiments of different backbone networks based on YOLOv8 demonstrate that the optimized GhostHGNetv2 achieves a remarkable balance between performance and computational efficiency. Specifically, GhostHGNetv2 attains 87.1% mAP@0.5 with only 6.8 GFLOPs, outperforming other computationally efficient networks such as GhostNet (79.3%) and Swin Transformer (80.5%). The model excels in detecting multiple categories, particularly showing outstanding performance in "person" (95.1%) and "fighting" (94.7%) detection tasks, highlighting its efficient detection capability in complex scenarios. In comparison, other models like Mobilenetv4 (22.5 GFLOPs) and Swin Transformer (79.1 GFLOPs), despite their superior performance in certain tasks, face limitations in practical deployment due to higher computational demands. In summary, the optimized GhostHGNetv2 not only achieves breakthroughs in accuracy but also maintains low computational resource consumption, demonstrating its superiority in real-time anomaly detection tasks. This makes it particularly suitable

4.5 **Ablation Study**

To systematically evaluate the contributions of each proposed module in HGO-YOLO, we conducted extensive ablation experiments under consistent training settings. As detailed in Table 4, the baseline model (b) achieved 84.4 mAP@0.5 at 33 FPS, with category-wise performance of 83.8 (fall), 92.4 (fight), 75.0 (smoke), and 86.5 (person).

In Table 4 we compared five different configurations of the model:

- H-YOLO(HGNetv2): When integrated solely into the backbone network, HGNetv2 improved mAP@0.5 by 1.5% (from 84.4 to 85.9), with particularly notable gains in the "fight" (+1.3%) and "person" (+7.0%) categories. This demonstrates its strong capability in multi-scale feature extraction.
- O-YOLO(OptiConvDetect): Upon the introduction of the OptiConvDetect detection head alone, the mAP@0.5 increased by 2.1% compared to YOLOv8's Decoupled-Head, while GFLOPs were reduced by 41%. This demonstrates that the combination of parameter sharing and PConv surpasses traditional designs in terms of both accuracy and efficiency.
- HG-YOLO(+GhostConv): Replacing standard convolutions with GhostConv in the HGNetv2 backbone reduced GFLOPs by 23.6% (from 6.9 to 6.8) and model size by 2.0% (from 4.9MB to 4.8MB), albeit with a marginal accuracy drop (mAP decreased by 0.2%). This highlights its effectiveness in minimizing computational costs and model complexity.
- HO-YOLO: The system uses HGNetv2 and the OptiConvDetect detection head. The FPS reached 52, and the mAP@0.5 was 86.0. This setup did very well in many categories. It performed especially well in the person category. This shows that the two improvements work well together to balance speed and accuracy.
- HGO-YOLO model: This architecture integrates the HGNetv2 backbone, GhostConv module, and OptiConvDetect head, achieving state-of-the-art performance at 56 FPS with 87.4 mAP@0.5. While demonstrating significant improvements across most categories, the model exhibited a slight 0.6% accuracy degradation (75.4% vs. baseline 76.0%) in smoke detection. This minor regression may be attributed to parameter sharing in OptiConvDetect potentially diluting feature representation for small targets like smoke.

**Tabel 3**: Comparison of Different Backbone Networks Experimental Results.

| Models | FLOPs(G) | mAP@0.5 | | | | | mAP@0.5:0.95 |
|---|---|---|---|---|---|---|---|
| | | fall | fight | smoke | person | ALL | |
| EfficientViT | 9.4 | 79.1 | 87.3 | 58.5 | 92.8 | 79.4 | 52.7 |
| VanillaNet | 96.7 | 78.8 | 86.2 | 61.1 | 93.5 | 79.9 | 54.0 |
| FasterNet | 10.7 | 78.1 | 88.9 | 63.5 | 94.9 | 81.4 | 54.4 |
| GhostNet | 5.0 | 79.9 | 88.5 | 54.6 | 94.3 | 79.3 | 52.5 |
| Mobilenetv4 | 22.5 | 80.8 | 90.3 | 61.5 | 93.2 | 81.4 | 54.5 |
| Swin Transformer | 79.1 | 78.8 | 89.2 | 60.5 | 93.6 | 80.5 | 54.7 |
| HGNetv2 | 6.9 | 82.7 | 93.7 | 73.5 | 93.5 | 85.9 | 61.5 |
| GhostHGNetv2 | 6.8 | 83.9 | 94.7 | 74.8 | 95.1 | 87.1 | 60.7 |

**Table 4**: Ablation study results, the baseline YOLOv8n model.

| | **Models** | | | **FPS** | **mAP@0.5** | | | | | **size** | |
|---|---|---|---|---|---|---|---|---|---|---|---|
| | H | G | O | | ALL | fall | fight | smoke | person | FLOPs(G) | #Param. |
| **b** | | | | 33 | 84.4 | 83.8 | 92.4 | 75.0 | 86.5 | 8.9 | 6.2MB |
| **1** | √ | | | 54 | 85.9 | 82.7 | 93.7 | 73.5 | 93.5 | 6.9 | 4.9MB |
| **2** | √ | √ | | 52 | 87.1 | 83.9 | 94.7 | 74.8 | 95.1 | 6.8 | 4.8MB |
| **3** | | | √ | 51 | 86.2 | 83.7 | 93.2 | 73.3 | 94.5 | 5.5 | 5.1MB |
| **4** | √ | | √ | 52 | 86.0 | 83.2 | 93.3 | 73.3 | 94.2 | 4.3 | 3.8MB |
| **5** | √ | √ | √ | 56 | 87.4 | 85.1 | 93.1 | 75.4 | 96.0 | 4.3 | 4.6MB |

4.6 **Analysis of Experimental Results at Different Scales**
we highlight HGO-YOLO's consistent improvements across scales (Table 5), such as HGO-YOLOn's 3.0% higher mAP@0.5 and 51.69% lower GFLOPs compared to YOLOv8n. Fig 6 now visually contrasts these gains, emphasizing our model's superior accuracy-efficiency balance across various model scales.

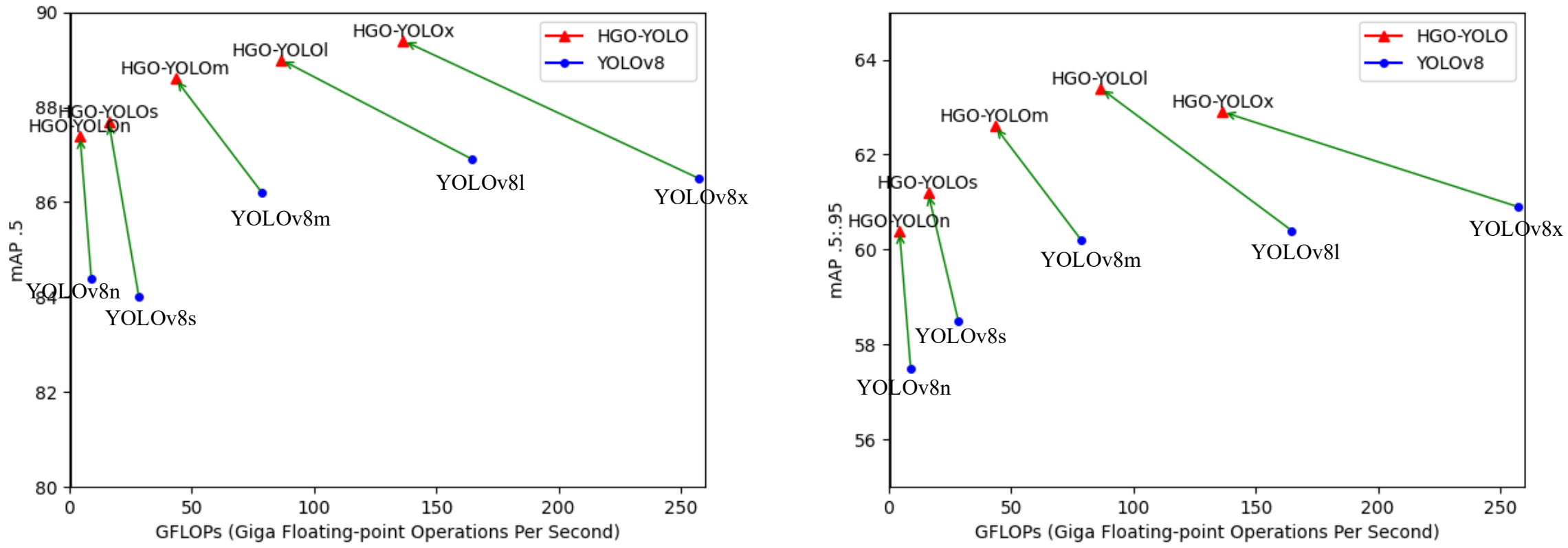

Fig 6: Compared with the baseline detection performance of HGO-YOLO.

**Table 5**: Comparing experimental results of different scales with YOLOv8 detection baseline.

| Model | #Param. | GFLOPs | FPS | Detection | | | | mAP@0.5 | | | |
|---|---|---|---|---|---|---|---|---|---|---|---|
| | | | | Precision | Recall | mAP@0.5 | mAP@0.5:0.95 | Fall | Fight | Smoke | Person |
| **YOLOv8n** | 6.2MB | 8.9 | 33 | 84 | 79,1 | 84.4 | 57,5 | 83.8 | 92.4 | 76 | 85.5 |
| **HGO-YOLOn** | 4.6MB | 4.3 | 56 | 87.1 | 81.1 | 87.4 | 60.4 | 85.1 | 93.1 | 75.4 | 96 |
| **improvement** | 1.6 | 4.6 | 23 | 3.1 | 2 | 3 | 2.9 | 1.3 | 0.7 | -0.6 | 10.5 |
| **YOLOv8s** | 22.5MB | 28.4 | 13 | 85 | 80.1 | 84 | 58.5 | 82.7 | 93.3 | 77.5 | 82.4 |
| **HGO-YOLOs** | 13,9 | 16 | 15 | 86.6 | 82.3 | 87.7 | 61.2 | 84.5 | 94.8 | 77 | 94.4 |
| **improvement** | 8.6 | 12.4 | 2 | 1.6 | 2.2 | 3.7 | 2.7 | 1.8 | 1.5 | -0.5 | 12 |
| **YOLOv8m** | 52MB | 78.7 | 6 | 86.7 | 80.5 | 86.2 | 60.2 | 85.3 | 93.1 | 80.8 | 85.5 |
| **HGO-YOLOm** | 30.5MB | 43.3 | 7 | 89 | 82.8 | 88.6 | 62.6 | 86.6 | 95 | 78.8 | 94.1 |
| **improvement** | 21.5 | 35.4 | 1 | 2.3 | 2.3 | 2.4 | 2.4 | 1.3 | 1.9 | -2 | 8.6 |
| **YOLOv8l** | 87.6MB | 164.8 | 3 | 86.6 | 79.8 | 86.9 | 60.4 | 85.2 | 95.1 | 81.2 | 85.9 |
| **HGO-YOLOl** | 49.5MB | 86.5 | 5 | 88.7 | 83.7 | 89 | 63.4 | 88 | 95.2 | 80 | 92.9 |
| **improvement** | 38.1 | 78.3 | 2 | 2.1 | 3.9 | 2.1 | 3 | 2.8 | 0.1 | -1.2 | 7 |
| **YOLOv8x** | 136.7MB | 257.4 | 2 | 84.3 | 81.8 | 86.5 | 60.9 | 85.3 | 94.5 | 81,5 | 84.9 |
| **HGO-YOLOx** | 77MB | 136.5 | 3 | 89.4 | 83.8 | 89.4 | 62.9 | 87.5 | 95.4 | 80.3 | 94.3 |
| **improvement** | 59.7 | 120.9 | 1 | 5.1 | 2· | 2.9 | 2 | 2.2 | 0.9 | -1.2 | 9.4 |

4.7 **Analysis of PConv and Conv Substitution Positions**
Table 6 compares the performance of four different replacement positions on the OptiConvDetect detector head. The results indicate that the PConv-Conv model achieved the highest overall accuracy at 87.4%, albeit with a relatively high computational cost. It notably performed well in detecting smoke anomalies with an accuracy of 75.4%. On the other hand, the Conv-PConv model, positioned differently, had slightly lower computational complexity and showed an advantage in the fall category. However, it experienced a 2.3% decrease in accuracy for the smoke category, resulting in a 0.5% drop in overall accuracy. The experiments conducted reveal that beginning with a conventional convolutional layer can result in the loss of intricate information during the feature extraction process. Transitioning to PConv can sometimes result in convolutional layers failing to recover or emphasize essential details from the original input, reducing accuracy in detecting small smoke targets. In the PConv-PConv experiment, while the "fall" category showed the best performance, the "smoke" category had the lowest accuracy. This is likely due to the exclusive use of PConv, which may limit the diversity of features and richness of information, preventing the model from fully capturing the data's complexity.

Although PConv excels in handling occlusions and extracting features from specific regions, relying on it alone can reduce the model's flexibility and adaptability across different tasks. On the other hand, the Conv-Conv model yielded the lowest overall accuracy, showing no clear advantage in any anomaly behavior. This is likely due to traditional convolutional layers treating all input features equally, without focusing on more critical parts of the input.

For tasks requiring high accuracy with manageable computational demands, the PConv-Conv model stands out as an optimal choice. It combines partial convolution (PConv) for initial feature extraction with traditional convolution (Conv) to refine and process features. PConv is effective in managing incomplete data or attention mechanisms, while the subsequent Conv layer enhances feature representation for better target detection. This hybrid approach improves the model's ability to detect and localize targets in complex scenes while maintaining computational efficiency.

**Table 6**: Comparison of the Improvement Effect of the Lightweight Detector Head at Different Positions.

| **OptiConvDetect** | **FLOPs(G)** | **mAP@0.5** | | | | |
|---|---|---|---|---|---|---|
| | | fall | fight | smoke | person | ALL |
| **Conv-Conv** | 4.3 | 83.8 | 93.5 | 73.9 | 94.1 | 86.3 |
| **Conv-PConv** | 4.2 | 85.7 | 94.5 | 73.1 | 94.2 | 86.9 |
| **PConv-Conv** | 4.3 | 85.1 | 93.1 | 75.4 | 96 | 87.4 |
| **PConv-PConv** | 4.1 | 86.2 | 93.3 | 73.0 | 94.4 | 86.7 |

### 4.8 **Comparison with Other Methods**

To comprehensively evaluate the performance of HGO-YOLO, we conducted a series of experiments on the dataset and detailed the detection results in Table 7. The experimental results show that HGO-YOLO achieves a mAP@0.5 of 87.4 while maintaining an excellent 56 FPS, leading in both metrics. Additionally, compared to RT-DETR, HGO-YOLO shows significant improvements in mAP@0.5 and detection speed, with a 6.2 mAP@0.5 increase and 7x FPS improvement. Compared to YOLOv5n, YOLOv6n, YOLOv7-tiny, and YOLOv8n, our model achieves higher mAP@0.5 scores of 2.8, 3.3, 4.2, and 3, respectively. In terms of detection speed, HGO-YOLO's 56 FPS is the highest. It is noteworthy that our model performs most accurately in detecting anomaly behaviors such as fall and fight.

When compared to SSD and Faster R-CNN, HGO-YOLO shows significant improvements in both accuracy and speed. SSD achieves a mAP@0.5 of 74.2 and runs at 10 FPS, while Faster R-CNN achieves a mAP@0.5 of 75.6 and runs at 8 FPS. Overall, HGO-YOLO effectively detects anomaly behaviors while maintaining a relatively fast detection speed.

**Table 7**: Comparison with other methods, the maximum values are highlighted in bold.

| Model | #Param. | FPS | mAP@0.5 | | | | | mAP@0.5:0.95 |
|---|---|---|---|---|---|---|---|---|
| | | | fall | fight | smoke | person | ALL | |
| SSD | 16.8M | 10 | 73.3 | 82.3 | 63.0 | 85.4 | 74.2 | 48.8 |
| Faster R-CNN | 33.5M | 8 | 76.3 | 85.4 | 66.2 | 85.5 | 75.6 | 50.2 |
| RT-DETR | 40.5M | 8 | 78.2 | 88.9 | 74.7 | 82.9 | 81.2 | 54.4 |
| YOLOv5n | 5.3M | 32 | 83.1 | 91.9 | 76.8 | 86.5 | 84.6 | 57.5 |
| YOLOv6n | 8.7M | 36 | 84.3 | 92.1 | 73.4 | 86.8 | 84.1 | 56.8 |
| YOLOv7-tiny | 12.3M | 27 | 82.8 | 90.3 | 74.5 | 85.2 | 83.2 | 56.9 |
| YOLOv8n | 6.3M | 33 | 83.8 | 92.4 | 76.0 | 85.5 | 84.4 | 57.5 |
| YOLOv9t | 6.1M | 10 | 84.5 | 92.5 | 76.9 | 87.3 | 85.1 | 58.2 |
| YOLOv10n | 5.8M | 41 | 84.0 | 92.0 | 72.3 | 91.9 | 85.0 | 56.7 |
| YOLOv11n | 5.3M | 16 | 84.6 | 91.5 | 77.2 | 86.7 | 85.0 | 57.9 |
| HGO-YOLO | 4.6M | 56 | 85.1 | 93.1 | 75.4 | 96.0 | 87.4 | 60.4 |

As shown in the Table 8, although EfficientDet-D0 has the lowest FLOPs, its mAP is 4.2% lower than that of HGO-YOLO and its FPS is also inferior; YOLOX-nano slightly surpasses HGO-YOLO in speed (63 vs. 56 FPS) but trails by 2.7% in mAP; EfficientViT-D1 achieves a compromise between speed and model size, yet its mAP still falls short of HGO-YOLO. Overall, HGO-YOLO attains a balanced performance—87.4% mAP@0.5 and 56 FPS—with only 4.3 GFLOPs and a 4.6 MB model size, clearly demonstrating its optimal trade-off among speed, accuracy, and resource consumption.

**Table 8**: Comparison of HGO-YOLO versus recent lightweight and anomaly-detection architectures.

| Model | mAP@0.5 (%) | FPS | GFLOPs | Model Size (MB) |
|---|---|---|---|---|
| EfficientViT-D0 | 83.2 | 42 | 2.5 | 5.2 |
| EfficientViT-D1 | 82.9 | 49 | 3.9 | 5.5 |
| YOLOX-nano | 84.7 | 63 | 3.1 | 4.0 |
| HGO-YOLO | 87.4 | 56 | 4.3 | 4.6 |

### 4.9 **Experiments with Different Loss Functions**

This paper investigates various IoU-based loss functions and evaluates their performance through experiments. The tested loss functions include DIoU loss[24], CIoU loss[25], MPDIoU loss[26], and Inner-CIoU loss[27]. As shown in Table 9, when using MPDIoU as the loss function for HGO-YOLO, the model achieved the highest mAP of 87.4%. This advantage stems from MPDIoU's comprehensive optimization of the geometric relationships of bounding boxes, especially showing higher localization stability in complex occlusion scenarios. To better illustrate the convergence of these loss functions during training, we present a comparison of the loss values for different functions across training cycles. As shown in Fig 7, the MPDIoU loss function converges the fastest and most steadily, while DIoU and CIoU converge more slowly, indicating that they may require more training epochs to reach optimal performance. These observations highlight the importance of selecting the appropriate loss function for different detection tasks. The advantages of MPDIoU in convergence speed and final mAP performance make it the most suitable choice for our anomaly detection task.

**Table 9**: Comparison of Various Losses

| Loss | DIoU | MPDIoU | CIoU | Inner-CIoU |
|---|---|---|---|---|
| mAP | 86.9 | 87.4 | 87.1 | 86.4 |

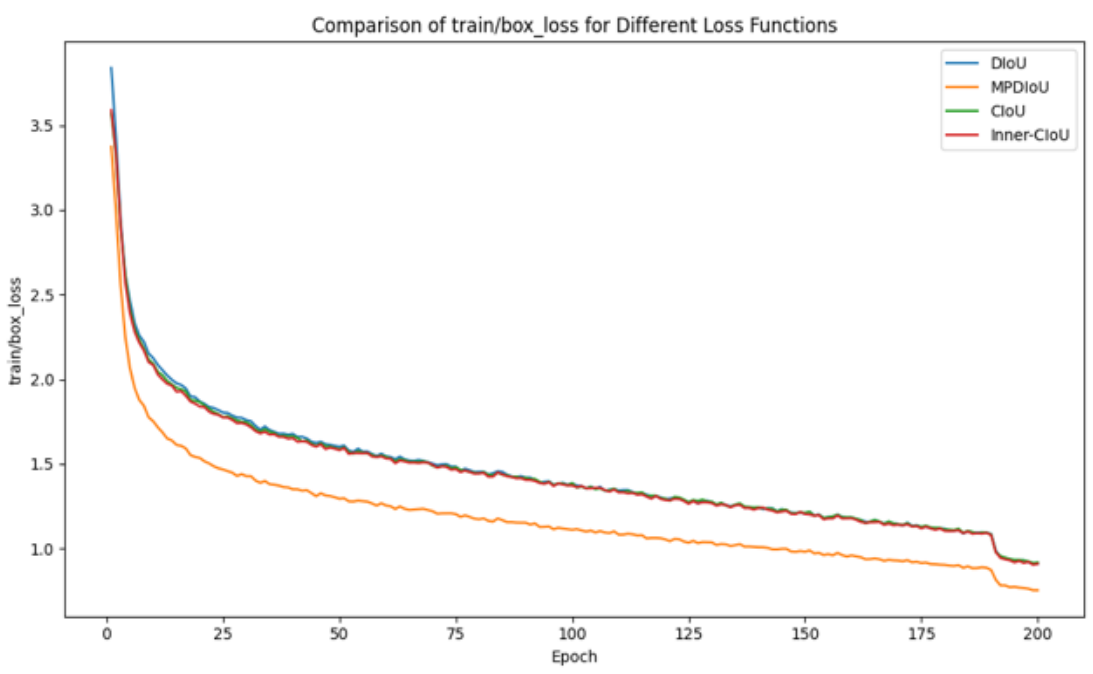

Fig 7: The comparison of loss convergence during training.

4.10 **Test**

To comprehensively evaluate the real-time detection performance and resource consumption of the improved model on different devices, we deployed YOLOv8 and HGO-YOLO on the Raspberry Pi 4 and NVIDIA Jetson Orin Nano platforms, and tested their frame rates (FPS), total CPU usage, memory consumption (RAM), and power consumption. The test input size was standardized to $480 \times 640$ and the results are shown in Table 10.

The experimental results show that HGO-YOLO significantly outperforms the baseline model in terms of computational efficiency and resource utilization. On the Raspberry Pi 4 platform, HGO-YOLO improves the frame rate from 1 FPS to 2 FPS. However, due to the hardware limitations of the Raspberry Pi, there is still room for further performance optimization. Future work can further reduce computational load through techniques such as model quantization and pruning. Additionally, HGO-YOLO reduces memory usage, power consumption, and CPU usage, effectively alleviating resource constraints. On the Jetson Orin Nano platform, HGO-YOLO achieves a frame rate of 42 FPS, a 16.7% improvement over YOLOv8, while also reducing memory usage, power consumption, and CPU usage. These improvements demonstrate that HGO-YOLO significantly enhances the operational efficiency of embedded devices while maintaining detection accuracy, making it particularly suitable for resource-constrained scenarios such as edge computing devices.

**Table 10**: Real-time detection performance and resource consumption of YOLOv8 and HGO-YOLO on different devices.

| Device | Model | Input Size | FPS | CPU(%) | RAM(MB) | Power(W) |
|---|---|---|---|---|---|---|
| Paspberry Pi 4 | YOLOv8 | $480 \times 640$ | 1 | 99% | 450 | 6.1 |
| | HGO-YOLO | $480 \times 640$ | 2 | 95% | 400 | 5.5 |
| Jetson Orin Nano | YOLOv8 | $480 \times 640$ | 36 | 19% | 1110 | 4.6 |
| | HGO-YOLO | $480 \times 640$ | 42 | 17% | 950 | 4.5 |

4.11 **Visualization**

To compare the performance of the HGO-YOLO model with the baseline model on the anomaly detection dataset, we conducted a series of visualizations.

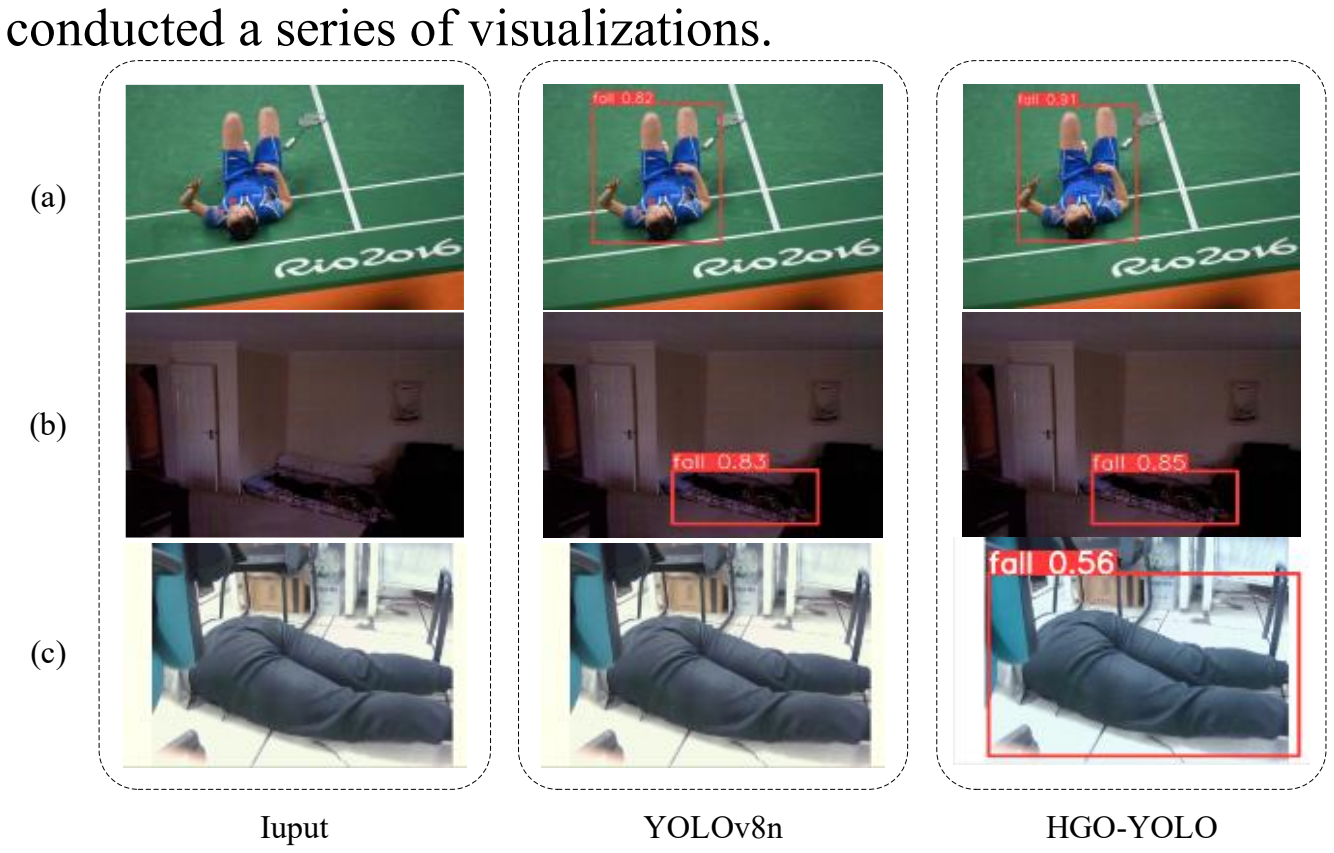

Fig 8: A comparison of detection performance for the fall category between YOLOv8n and HGO-YOLO.

As shown in Fig 8, we present a comparison of prediction results for the fall category between YOLOv8n, and HGO-YOLO based on the same validation input data. In a dimly lit scenario (b), YOLOv8n detects the fall (confidence: 0.83), while HGO-YOLO correctly identifies the target (confidence: 0.85). In a real-world occlusion scenario (c), HGO-YOLO avoids missed detection caused by partial occlusion, demonstrating robustness in complex environments. The experiments show that even in challenging scenarios such as low-light conditions and severe occlusion, HGO-YOLO achieves significantly higher detection confidence for complex poses compared to YOLOv8n. These findings confirm that HGO-YOLO can provide more reliable performance for real-time detection systems.

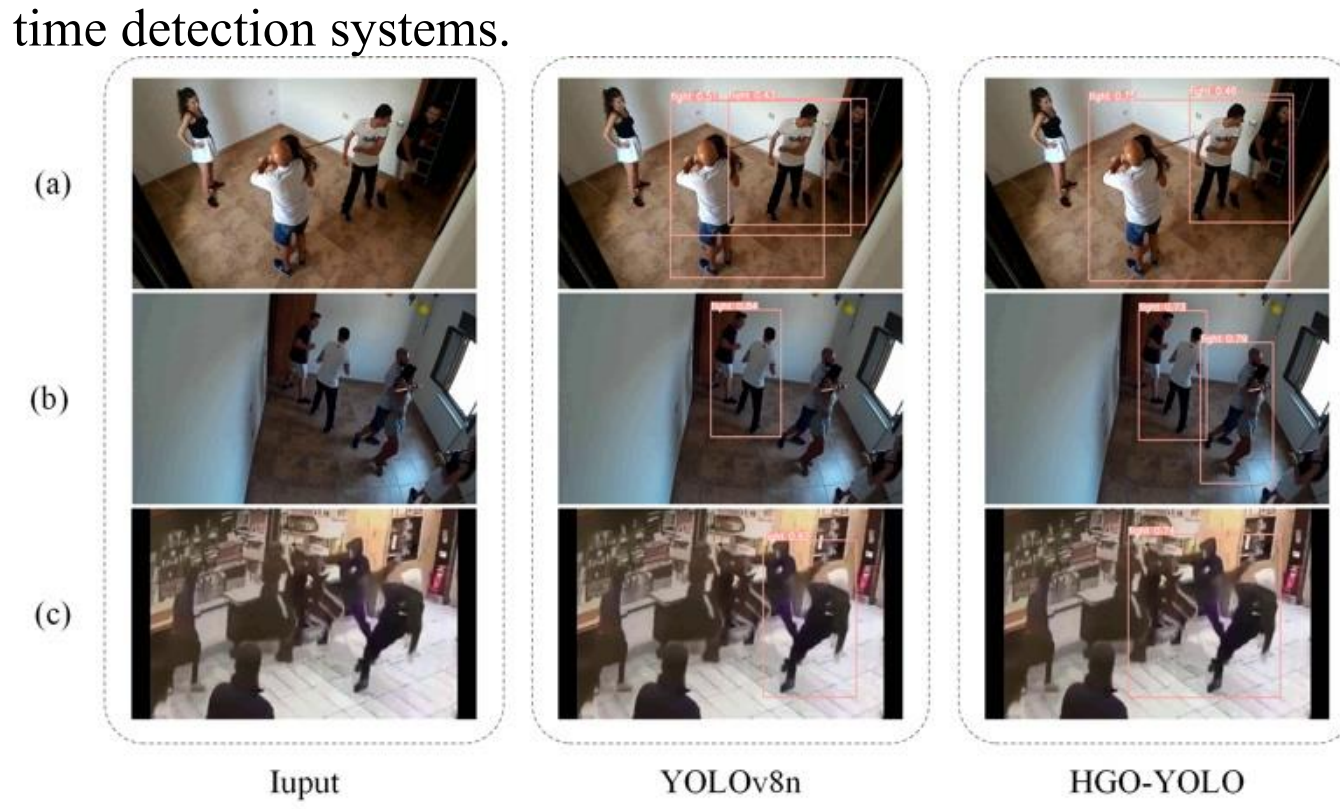

Fig 9: A comparison of the detection performance for the fight category between YOLOv8n and HGO-YOLO.

Fig 9 highlights the advantages of HGO-YOLO in crowded and occluded fight scenarios. In an indoor environment (Fig. 9a), YOLOv8n exhibits lower detection confidence (0.43–0.51) and fails to clearly distinguish details between objects or individuals, whereas HGO-YOLO maintains stable performance (confidence: 0.46–0.75). In a dimly lit, confined indoor scene (Fig. 9b), although YOLOv8n achieves a detection confidence of 0.84, it suffers from missed detections, while HGO-YOLO remains stable (confidence: 0.73–0.79). For crowded indoor fights (Fig. 9c), our model achieves a confidence score of 0.74, with bounding boxes that more closely match the actual target's shape and position, indicating improved localization accuracy and highlighting its anti-interference capability in dense scenarios.

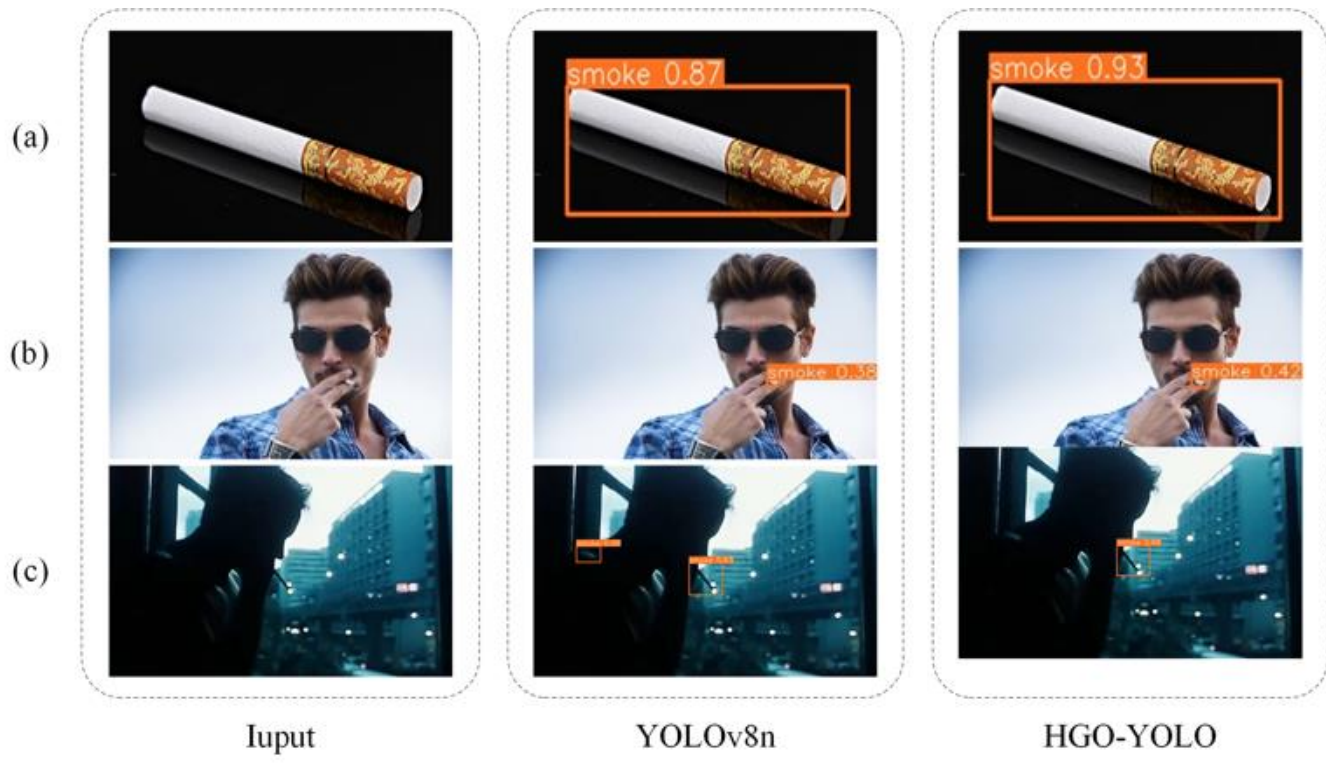

Fig 10: Smoke category detection performance comparison between YOLOv8n and HGO-YOLO.

Fig 10 presents a comparison of detection results for the smoking category. YOLOv8n exhibits false detections for small targets, whereas HGO-YOLO performs exceptionally well in low-contrast nighttime scenarios (Fig. 10c). YOLOv8n tends to misclassify reflections as smoke in such scenes, while our model suppresses such errors while maintaining sensitivity to real smoking regions. In summary, HGO-YOLO demonstrates higher confidence in smoking detection and localization compared to YOLOv8n.

## 5. Conclusion

This study proposes the HGO-YOLO model, which marks a significant advance in real-time anomaly detection, achieving 87.4 % mAP@0.5 at 56 FPS—substantially outperforming the YOLOv8 baseline's 84.4 % mAP@0.5 at 33 FPS. Although the model exhibits outstanding detection capability in most scenarios, it still shows a 0.6 % accuracy gap when handling tiny objects such as smoking events. This limitation mainly stems from the parameter-sharing strategy adopted in the OptiConvDetect module, which can cause the loss of fine-grained spatial features crucial for small-object detection. To resolve this issue, we plan to incorporate channel- and spatial-attention mechanisms to enhance feature discrimination, optimize the feature-pyramid structure to reduce down-sampling in deeper layers, and apply mixed-precision quantization together with structured pruning for model compression[28, 29] These improvements are expected to further boost performance while maintaining edge-computing efficiency[30], thereby closing the small-object accuracy gap and potentially setting new performance benchmarks[31, 32]. Moreover, Grad-CAM heatmaps can be introduced to satisfy current explainable-AI requirements[33].

From an ethical standpoint, deploying HGO-YOLO in surveillance scenarios requires heightened attention to personal-privacy protection. Because the system processes video data that may contain sensitive information, strict adherence to data anonymization[34], secure storage, and usage-scope limitations[35] is essential. To ensure fairness and transparency, potential biases in both the dataset and model behavior must be regularly assessed. In future work, we will explore privacy-preserving techniques such as federated learning to further enhance data-security safeguards.